# Continental-scale land cover mapping at 10 m resolution over Europe (ELC10)

Zander S. Venter[1]; Markus A.K. Sydenham[1]

[1]Terrestrial Ecology Section, Norwegian Institute for Nature Research - NINA, Oslo, Norway

## Abstract

Land cover maps are important tools for quantifying the human footprint on the environment and facilitate reporting and accounting to international agreements addressing the sustainable development goals. Widely used European land cover maps such as CORINE (Coordination of Information on the Environment) are produced at medium spatial resolutions (100 m) and rely on diverse data with complex workflows requiring significant institutional capacity. We present a high resolution (10 m) land cover map (ELC10) of Europe based on a satellite-driven machine learning workflow that is annually updatable. A Random Forest classification model was trained on 70K ground-truth points from the LUCAS (Land Use/Cover Area frame Survey) dataset. Within the Google Earth Engine cloud computing environment, the ELC10 map can be generated from approx. 700 TB of Sentinel imagery within approx. 4 days from a single research user account. The map achieved an overall accuracy of 90% across 8 land cover classes and could account for statistical unit land cover proportions within 3.9% ($R^2$ = 0.83) of the actual value. These accuracies are higher than that of CORINE (100 m) and other 10-m land cover maps including S2GLC and FROM-GLC10. Spectro-temporal metrics that capture the phenology of land cover classes were most important in producing high mapping accuracies. We found that atmospheric correction of Sentinel-2 and speckle filtering of Sentinel-1 imagery had minimal effect on enhancing classification accuracy (< 1%). However, combining optical and radar imagery increased accuracy by 3% compared to Sentinel-2 alone and by 10% compared to Sentinel-1 alone. The addition of auxiliary data (terrain, climate and nighttime lights) increased accuracy by an additional 2%. The conversion of LUCAS points into homogenous polygons under the Copernicus module increased accuracy by <1%, revealing that Random Forests are robust against contaminated training data. Furthermore, the model requires very little training data to achieve moderate accuracies - the difference between 5K and 50K LUCAS points is only 3% (86 vs 89%). This implies that significantly less resources are necessary for making in-situ survey data (such as LUCAS) suitable for satellite-based land cover classification. At 10-m resolution, the ELC10 map can distinguish detailed landscape features like hedgerows and gardens, and therefore holds potential for aerial statistics at the city borough level and monitoring property-level environmental interventions (e.g. tree planting). Due to the reliance on purely satellite-based input data, the ELC10 map can be continuously updated independent of any country-specific geographic datasets.

## 1. Introduction

Satellite-based remote sensing of land use and land cover has afforded dynamic monitoring and quantitative analysis of the human footprint on the biosphere (Chang et al., 2018). This is important because land cover change is a significant driver of the global carbon cycle, energy balance and biodiversity changes (Foley et al., 2005; Maxwell et al., 2016) which are processes of existential consequence. Land cover maps are often the primary inputs into accounting frameworks that attempt to monitor countries' efforts towards addressing the Sustainable Development Goals (SDG) (Holloway and Mengersen, 2018). For instance, land cover maps are often used to set targets and indicators for meeting SDG 2 of zero hunger and SDG 15 to monitor efforts to reduce natural habitat loss (e.g. deforestation alerts). Ecosystem service models and accounts also rely on land cover data as input (de Araujo Barbosa et al., 2015) and land cover maps are thereby important for the valuation and conservation of important ecosystems. In light of global climate change and a rapidly developing world, an increasing number of applications, such as precision agriculture, wildlife habitat management, urban planning, and renewable energy installations, require higher resolution and frequently updated land cover maps.

The advent of cloud computing platforms like Google Earth Engine (Gorelick et al., 2017) has led to significant advances in the ability to map land surface changes over time (Mahdianpari et al., 2020; Tamiminia et al., 2020). This is both due to the enhanced computing power and the availability of dense time series data from medium to high resolution sensors like Sentinel-2 (Drusch et al., 2012). The transition to time series imagery allows one to capture the seasonal and phenological components of land cover classes that would otherwise be missed with single time-slice imagery. The application of such spectro-temporal metrics to mapping forest (Potapov et al., 2015) and other land cover types (Azzari and Lobell, 2017) have shown increased classification accuracies. In addition, the ability to adopt machine learning algorithms in cloud computing environments has further enhanced the precision of land cover mapping (Holloway and Mengersen, 2018).

The CORINE (Coordination of Information on the Environment) land cover map of Europe (Büttner, 2014) is perhaps the most widely used land cover product for area statistics and research (Bielecka and Jenerowicz, 2019). The CORINE map currently requires significant institutional capacity and coordination from the European Economic Area members, the Eionet network, and the Copernicus programme. For instance the 2012 product involved 39 countries, a diversity of country-specific topographic and remote sensing datasets, and took two years to complete. To ease the manual workload, the wealth of data from the Copernicus Sentinel sensors has been somewhat integrated into the CORINE mapping workflow, and has also led to the development of Copernicus Land cover services high spatial resolution maps (https://land.copernicus.eu/pan-european/high-resolution-layers). Recently, Sentinel-2 data has been used to create a 10m pan-European land cover/use map (S2GLC) for cairca 2017 (http://s2glc.cbk.waw.pl/) (Malinowski et al., 2020). This is a meaningful advancement on previous pan-European mapping efforts however, the methodology behind S2GLC involves a land cover reference dataset and some post-processing steps that are not open-source or easily reproducible. Pflugmacher et al., (2019) recently developed an independent, research-driven approach to pan-European land cover mapping with Landsat data at 30 m for cairca 2015. This compares favourably with the CORINE map, is reproducible and does not require harmonising and collating country-specific datasets from different European member states. Nevertheless, there remains potential for a similar open-source approach that leverages both Sentinel-2 optical and Sentinel-1 radar sensor data to map land cover at 10-m resolution (Phiri et al., 2020).

Land cover maps made with open data policies and open science principles can have transfer value to other areas of the globe (Chaves et al., 2020) particularly when pre- and post-processing decisions are made transparent. Like the European maps mentioned above, the studies documenting continental land cover classifications at 30- or 10-m resolution for Africa (Li et al., 2020; Midekisa et al., 2017), North America (Zhang and Roy, 2017) and Australia (Calderón-Loor et al., 2021) have not communicated methodological lessons or published source code. The same is true for global land cover products such as the Landsat-based GLOBLAND30 (Jun et al., 2014) or Sentinel-based FROM-GLC10 (Gong et al., 2019). This makes it difficult to draw generalizable conclusions that benefit the remote sensing and land cover mapping community at large. Specifically, it is not clear how satellite and reference data pre-processing decisions affect the accuracy of land cover classifications at this scale. Such decisions may concern the atmospheric correction of optical imagery (Sentinel-2), the speckle filtering of radar imagery (Sentinel-1), or the fusion of optical and radar data within one classification model. When trying to classify land cover over very broad environmental gradients where spectral signatures vary substantially within a given land cover class, one may also decide to include auxiliary variables to increase model accuracy ((Pflugmacher et al., 2019). Such decisions have trade-offs between computational efficiency and classification accuracy which are important to quantify when operationalizing land cover classification at continental scales.

Another important point of consideration in operational land cover classification, is the collection and cleaning of reference data ("ground-truth") that are used to train a classification model. The quality, quantity and representativity of reference data can have significant effects on the accuracy and consequent utility of a land cover map (Chaves et al., 2020). In Europe, the Land Use/Cover Area frame Survey (LUCAS) dataset consists of in-situ land cover data collected over a grid of point locations over Europe (d'Andrimont et al., 2020a). However, when aligning satellite pixels data with LUCAS grid points, the geolocation uncertainty in both datasets can lead to mislabelled training data for land cover classification. To make LUCAS data suitable for earth observation, EUROSTAT introduced a new module (i.e. the Copernicus module) to the LUCAS survey in 2018 (d'Andrimont et al., 2020b). The Copernicus module has quality-assured and transformed 58 428 of the LUCAS points into polygons of homogeneous land cover that are suitable for earth observation purposes. Given that (Weigand et al., 2020) have shown that intersecting Sentinel pixels with LUCAS grid points already yields accurate land cover classifications, it remains to be seen how the inclusion of the Copernicus LUCAS polygons improves classification accuracy. Furthermore, previous attempts to integrate LUCAS data with remote sensing for land cover classification (Close et al., 2018; Pflugmacher et al., 2019; Weigand et al., 2020) have not fully assessed the trade-off between reference sample size, model accuracy and the spatial distribution of prediction uncertainty. This information is important for planning future ground-truth data collection missions and remote sensing integrations.

Here we aim to build upon previous efforts to generate a 10-m Sentinel-based pan-European land cover map (ELC10) for 2018 using a reproducible and open-source machine learning workflow. In doing so we aim to explicitly test the effect of several pre-model data processing decisions that are often overlooked. Concerning satellite data processing, these include the effect of (1) Sentinel-2 atmospheric correction; (2) Sentinel-1 speckle filtering; (3) fusion of optical and radar data; (4) addition of auxiliary predictor variables. Concerning land cover reference data, we aim to test the effect of (5) quality-checking reference points through the use of the LUCAS Copernicus module, and (6) the effect of decreasing reference sample size. Finally, we compare ELC10 to existing land cover maps both in terms of accuracy and utility for area statistics accounting.

## 2. Methods

### 2.1. Study area

We defined the scope of our study area to include all of Europe from 10°W to 30°E longitude and 35°N to 71°N latitude, except for Iceland, Turkey, Malta and Cyprus (Figure 1). This area is similar to the CORINE Land Cover product produced by the Copernicus Land Monitoring Service covering the European Economic Area 39 countries and approximately 5.8 million square kilometres. Europe covers a wide range of climatic and ecological gradients primarily explained by the North-South latitudinal gradient (Condé et al., 2002). Southern regions are arid warmer climates supporting a diverse range of Mediterranean vegetation. Northern regions are mesic, cooler climates characteristic of Boreal and Atlantic zones with shorter growing seasons and lower population densities leading to forest-dominated landscapes. Europe has a significant anthropogenic footprint with 40% of the land covered by agriculture, including semi-natural grasslands.

## 2.2. Land cover reference data

LUCAS is a European Union initiative to gather in-situ ground-truth data on land cover over 27 member states and is updated every three years (Gallego and Delincé, 2010). By definition it excludes Norway, Switzerland, Liechtenstein, and the non-EU Balkan states. Each iteration includes visiting a sub-sample of the 1 090 863 geo-referenced points within the LUCAS 2-km point grid. Under the 2018 LUCAS Copernicus module, 58 428 of the point locations have been quality assured and transformed into polygons of homogenous land cover specifically tailored for earth observation (Fig. 2). The polygons are approximately 0.5 ha in size and are therefore (by design) large enough so that at least one Sentinel 10 x 10 m pixel is contained fully within them with some space for registration error. We used the collated and cleaned Copernicus Module polygon dataset (n = 53476) provided by (d'Andrimont et al., 2020b). The Copernicus Module polygons (hereafter referred to as LUCAS polygons) were used as the core of our reference sample for land cover classification. The top-level of the LUCAS land cover typology was used in the present analysis including artificial land, cropland, woodland, shrubland, grassland, bare land, wetland, and water (Table 1).

After establishing baseline land cover proportions using the CORINE land cover dataset (re-coded to our typology) as reference (Büttner, 2014), we found that the distribution of the LUCAS polygons were biased toward cropland and woodland land cover classes (Fig. S1). Consequently, there were very few LUCAS polygons for water, wetland, bare land and artificial land classes (Fig. S1). We therefore performed a bias correction of the reference sample (Fig. 2) by using the harmonized LUCAS grid point (hereafter LUCAS points) data (d'Andrimont et al., 2020a) to supplement the LUCAS polygon dataset so that the overall reference sample was representative of the CORINE proportions. Although the LUCAS points have not been transformed into polygons, they are still appropriate for earth observation applications (Pflugmacher et al., 2019) after applying certain quality control procedures. We employed the metadata filtering (Fig. 2) outlined in (Weigand et al., 2020) to filter out points where the land cover parcel area was < 0.5 ha, or covered < 50% of the parcel. As in Pflugmacher et al., (2019) we also excluded classes with potential thematic and spectral ambiguity including linear artificial features (LUCAS LC1 code A22), other artificial areas (A39), temporary grasslands (B55), spontaneously re-vegetated surfaces (E30) and other bare land (F40). This resulted in 282 854 labelled point locations available to supplement the LUCAS polygon sample. Of these, 18 009 LUCAS points were selected following an outlier ranking procedure to remove mislabeled or contaminated LUCAS points.

The outlier ranking procedure involved extracting Sentinel-2 data (see section 2.3. for details) for pixels intersecting LUCAS points. These were fed into a Random Forest (RF) classification model (see section 2.5 for details) which was used to calculate classification uncertainty for each LUCAS point. The RF model iteratively selects a random subset of data to generate decision trees which are validated against the withheld data. During each iteration the model generates votes for the most likely class label. We extracted the fraction of votes for the correct land cover class at each LUCAS point after bootstrapping the RF procedure 100 times. We acknowledge that this bootstrapping of the RF model itself may not be necessary, however, it may smooth over any artifacts introduced from the internal bootstrapping of a single RF model. LUCAS points with a high fraction of votes (close to 1) can be considered as archetypal instances of the given land cover class, whereas those with a low fraction of votes (close to 0) are considered as mis-labelled or spectrally contaminated. We ranked the LUCAS points by their fraction of correct votes, and selected the topmost points for each land cover class to supplement the LUCAS polygons so that the final land cover proportions matched that of the CORINE dataset. The number of supplemental LUCAS points needed (n = 18 009) was determined relative to the most abundant LUCAS polygon class (cropland in Fig. S1).

## 2.3. Sentinel spectro-temporal features

All remote sensing analyses were conducted in the Google Earth Engine cloud computing platform for geospatial analysis (Gorelick et al., 2017). We processed all Sentinel-2 optical and Sentinel-1 synthetic aperture radar (SAR) scenes over Europe during 2018. This amounts to a total of 239 818 satellite scenes which would typically require approx. 700 TB storage space if not for Google Earth Engine and cloud computation. The Sentinel satellite data were used to derive spectro-temporal features as predictor variables in our land cover classification model. Spectro-temporal features have been used to capture both the spectral and temporal (e.g. phenology or crop cycle) characteristics of land cover classes and offer enhanced model prediction accuracy compared to single time-point image classification (Griffiths et al., 2019; Pflugmacher et al., 2019). To generate model training data, spectro-temporal metrics were extracted for Sentinel pixels intersecting the LUCAS points, or the centroids of the LUCAS polygons.

Sentinel-2 images for both Top of Atmosphere (TOA; Level 1C) and Surface Reflectance (SR; Level-2A) were used to test the effect of atmospheric correction on classification accuracies (Q1 in Fig. 2). The scenes were first filtered for those with less than 60% cloud cover (129 839 removed of 280 420 scenes) using the "CLOUDY_PIXEL_PERCENTAGE" scene metadata field. We then performed a pixel-wise cloud masking procedure using the cloud probability score produced by the S2cloudless algorithm (Zupanc, 2020). S2cloudless is a machine learning-based algorithm and is part of the latest generation of cloud detection algorithms for optical remote sensing images. After visually inspecting the cloud masking results across a range of Sentinel-2 scenes, we settled on a cloud probability threshold of 40% for our masking procedure. After cloud masking and mosaicing two year's worth of Sentinel-2 scenes, the cloud-free pixel availability ranged from less than 10 to over 100 pixels over the study area (Fig. 1b).

Using the cloud-masked Sentinel-2 imagery we derived the median mosaic of all spectral bands. In addition we calculated the following spectral indices for each cloud-masked scene: normalized difference vegetation index (Tucker, 1979), normalized burn ratio (García and Caselles, 1991), normalized difference built index (Zha et al., 2003) and normalized difference snow index (Nolin and Liang, 2000). For each spectral index we used the temporal resolution to calculate the 5th, 25th, 50th, 75th and 95th percentile mosaics as well as the standard deviation, kurtosis and skewness across the two-year time stack of imagery. We derived the median NDVI values for summer (Jun-Aug), winter (Dec-Feb), spring (Mar-May), and fall (Sep-Nov). The spectro-temporal metrics described above have been extensively used to map land cover and land use changes with optical remote sensing (Gómez et al., 2016). Finally, several studies have found that textural image features (i.e. defining pixel values from those of their neighborhood) for Sentinel-2 imagery significantly enhanced land cover classification accuracy (Khatami et al., 2016; Weigand et al., 2020). Therefore, we calculated the standard deviation of median NDVI within a 6 x 6 pixel moving window.

Sentinel-1 SAR Ground Range Detected data have been pre-processed by Google Earth Engine, including thermal noise removal, radiometric calibration and terrain correction using global digital elevation models. Sentinel-1 scenes were filtered for interferometric wide swath and a resolution of 10 m to suit our land cover classification purposes. We performed angular-based radiometric slope correction using the methods outlined in (Vollrath et al., 2020). SAR data can contain substantial speckle and backscatter noise which is important to address particularly when performing pixel-based image classification. We applied a Lee-sigma speckle filter (Lee et al., 2008) to the Sentinel-1 imagery to test the effect on classification accuracy (Q2 Fig. 2). Following pre-processing, we calculated median and standard deviation mosaics for the time stacks of imagery including the single co-polarized, vertical transmit/vertical receive (VV) band and the cross cross-polarized, vertical transmit/horizontal receive (VH) band, as well as the ratio between them (VV/VH).

## 2.4. Auxiliary features

A challenge with classifying regional-scale land cover is that models relying on spectral responses alone may be limited by the fact that land cover characteristics can change drastically between climate and vegetation zones. For example, a grassland in the meditterrainean will have very different spectro-temporal signatures to a grassland in the boreal zone. Previous regional land cover classification efforts have dealt with this by either (1) splitting the area up into many small parts and running multiple classification models (Zhang and Roy, 2017), or (2) including environmental covariates that help the model explain the regional variation in land cover characteristics (Brown et al., 2020; Pflugmacher et al., 2019). We tested the latter approach (Q4 in Fig. 2) by including a range of environmental auxiliary covariates into our classification model.

Auxiliary variables included elevation data from the Shuttle Radar Topography Mission (SRTM) digital elevation dataset (Farr and Kobrick, 2000) at 30m resolution which covers up to 60° North. For higher latitudes we used the 30 Arc-second elevation data from the United States Geological Survey (GTOPO30). Climate data were derived from the ERA5 fifth generation ECMWF atmospheric reanalysis of the global climate (Copernicus Climate Change Service, 2017). We used it to calculate 10-year (2010-present) average and standard deviation in monthly precipitation and temperature at 25 km resolution. Finally, we also included data on nighttime light sources at approx. 500m spatial resolution. This was intended to assist the model in differentiating from artificial surfaces and bare ground in alpine areas. A median 2018 radiance composite image from the Visible Infrared Imaging Radiometer Suite (VIIRS) Day/Night Band (DNB), provided by the Earth Observation Group, Payne Institute, was used (Mills et al., 2013).

## 2.5. Classification models and accuracy assessment

The land cover classification model evaluation and tuning were conducted in R with the 'randomForest' and 'caret' packages (R Core Team, 2019), while the final model inference over Europe was conducted in Google Earth Engine using equivalent model parameters. We chose an ensemble learning method namely the Random Forest (RF) classification model. RF deals well with large and noisy input data, accounts for non-linear relationships between explanatory and response variables, and is robust against overfitting (Breiman, 2001). A recent review of land cover classification literature found that the RF algorithm has the highest accuracy level in comparison to the other classifiers adopted (Talukdar et al., 2020). Classification accuracies were determined using internal randomized cross-validation procedures where error rates are determined from mean prediction error on each training sample $x_i$, using only the trees that did not have $x_i$ in their bootstrap sample (i.e. out-of-bag; Lyons et al., 2018). Predicted and observed land cover classes are used to build a confusion matrix from which one derives overall accuracy (OA), user's accuracy (UA), and producer's accuracy (PA). See (Stehman and Foody, 2019) for details.

A series of RF models were run at each step in the pre-processing tests (Fig. 2) in order to assess the effect of pre-processing decisions on classification accuracy. With each consecutive step, we chose the pre-processing option that yielded the highest accuracy to generate the data for the subsequent step. The final pre-processing sequence that led to the final RF model data are indicated by the underlined decisions in Fig. 2. When testing the effect of reference sample size (Q6 in Fig. 2), we iteratively removed 5% of the training dataset and assessed model performance. All 71 485 LUCAS locations (polygons and points) were used to train the final RF model. At this stage we performed recursive feature elimination which is a process akin to backward stepwise regression that prevents overfitting and reduces unnecessary computational load (Guyon et al., 2002). Recursive feature elimination produces a model with the maximum number of features and iteratively removes the weaker variables until a specified number of features is reached. In our case this was 15 features. The top predictor variables were selected based on the variable importance ranking using both mean decrease accuracy and mean decrease Gini coefficient scores (Hong Han et al., 2016). Finally, we also tuned the RF hyperparameters by iterating over a series of *ntree* (50 to 500 in 25 tree intervals) and *mtry* (1 to 10) and found the optimal (based on lowest model error rate) combination of settings to include an *ntree* of 100 and *mtry* set to the square root of the number of covariates (3.8).

Part of enhancing the usability of land cover maps is quantifying the spatial distribution of classification uncertainty. There are methods to derive pixel-based and sample-based uncertainty estimates that are spatially-explicit (Khatami et al., 2016; Tinkham et al., 2014; Venter et al., 2020). We adopt a sample-based uncertainty estimate by dividing the study area into 100-km equal-area grid squares defined by the EEA reference grid. For each grid cell we use our final trained RF model to make predictions against the LUCAS reference data within and build a confusion matrix to derive overall accuracy for the grid cell in question. We acknowledge that making predictions over reference samples that were included in model training is likely to inflate accuracy estimates. However, in this case we are interested in obtaining the relative distribution of accuracy over the study region to give insight into class non-separability and map reliability over space.

## 2.6. Comparison with other land cover maps

We compared our land cover product with two other global land cover products including CORINE (Büttner, 2014), and FROM-GLC10 (Gong et al., 2019), and two other European land cover maps including the map created by (Pflugmacher et al., 2019) and S2GLC (Malinowski et al., 2020). The CORINE map has been updated for 2018 at 100m resolution by the Copernicus Land Management Service and is widely used for aerial statistics and accounting. FROM-GLC10 is a global map produced with Sentinel satellite data at 10-m resolution. The S2GLC (Sentinel-2 Global Land Cover) map has been produced over Europe during 2017 using Sentinel 2 data at 10m resolution. The (Pflugmacher et al., 2019) map was produced for 2015 using Landsat data at 30m resolution. All land cover typologies were converted to the LUCAS typology used in our analysis for purposes of comparison (Table S1). The same accuracy assessment protocols described above were used to assess the accuracy of these maps using the same validation dataset (completely withheld from the training of our model).

Apart from assessing the classification accuracy, we tested the utility of the maps for calculating aerial land cover statistics over spatial units defined for the European Union by the nomenclature of territorial units (NUTS). We used NUTS level 2 basic regions which include population sizes between 0.8 and 3 million and are used for the application of regional policies. Area proportions for each land cover class and map product, including ELC10, were calculated for each of the NUTS polygons. Within each NUTS polygon we also calculated area proportions using the original LUCAS survey dataset. We regressed the mapped area proportions on the area proportions estimated from the LUCAS sample to assess the land cover map's utility for land cover accounting. Although the statistics derived from LUCAS dataset also have uncertainty associated with them, they are considered the only harmonized dataset for area statistics in Europe and were therefore used as the benchmark with which we compared the land cover maps.

# 3. Results

## 3.1. Effects of satellite data pre-processing

The pre-processing of Sentinel optical and radar imagery had very little effect on the overall classification accuracy (Figs. 3a & b). Specifically, atmospheric correction of Sentinel-2 and speckle filtering of Sentinel-1 imagery enhanced classification accuracy by less than 1% compared to models with TOA and non-speckle filtered imagery, respectively. This marginal difference was true for all class-specific accuracies (Fig. S2). However, the fusion of Sentinel-1 and Sentinel-2 data within a single model increased accuracy by 3% compared to Sentinel-2 alone and by 10% compared to Sentinel-1 alone (Fig. 3c). Class-specific accuracies reveal that models with Sentinel-1 data alone perform particularly badly when predicting wetland, shrubland and bare land classes (Fig. S2c). In these instances, fusing both optical and radar data increases accuracy by up to 30% compared to Sentinel-1 data alone. The addition of auxiliary data (terrain, climate and nighttime lights) increased accuracy by an additional 2% compared to a model with Sentinel data alone (Fig. 3d). Auxiliary data had the largest benefits for bare land and shrubland classes (Fig. S2d).

## 3.2. Effects of reference data pre-processing

The first test of reference data pre-processing was a test of quality checking and cleaning the LUCAS data via the conversion of LUCAS points into homogenous polygons under the Copernicus module (Fig. 2). Extracting the satellite data at LUCAS points versus the centroids of homogenous LUCAS polygons increased accuracy by less than 1% (Fig. 3e). This marginal effect was evident for all class-specific accuracy scores (Fig. S2e). The second test related to reference data involved the iterative depletion of the sample size. The relationship between sample size and overall accuracy appears to follow an exponential plateau curve (Fig. 4). The benefit to model accuracy gained by increasing sample size depletes rapidly so that, for example, when one increases from 5K to 20K points, accuracy increases by 0.15% per 1K points added, while when one increases from 55K to 70K points, accuracy increases by 0.015% per 1K points. Therefore, the difference between 5K and 50K LUCAS points is only 3% (86 vs 89%; Fig. 4). The same pattern is evident for class-specific accuracies. However, it is important to note that the variance in accuracy from the bootstrapped RF classifications increased as the number of training samples decreased.

## 3.3. ELC10 final accuracy assessment

The final RF classification model produced an overall accuracy of 90.2% across 8 land cover classes (Table 2). The class-specific user's accuracy (UA; errors of commission) describes the reliability of the map and informs the user how well the map represents what is really on the ground. UA exhibited a wide range from 75% for shrubland to 96.4% for woodland. The relative decrease in prediction accuracy over shrubland classes is evident in the spatial distribution of model errors (Fig. 5). The majority of error (accuracies below 80%) was distributed over southern Europe where shrubland dominates (Fig. 1a). Conversely, model accuracies were highest (above 90%) over the interior of Europe (Fig. 5) where cropland and woodland dominate (Fig. 1a). Shrubland was most often confused with grassland and woodland probably due to the spectral similarity across a gradient of woody plant cover. Similarly, cropland was most often confused with grasslamd probably due to the temporal similarity in spectral signatures between mowed pastures and ploughed fields.

Sentinel optical variables were the two most important covariates in the final RF model (Fig. 6). The first and fifth most important variables were the 25th percentile of NDVI and standard deviation in NBR over time, respectively. These metrics both capture the temporal dynamics of spectral responses that are important in distinguishing land cover classes such as cropland and grassland. The Sentinel 1 VH band also exhibited a relatively high importance score. Of the auxiliary variables, nighttime light intensity and temperature were the most important variables.

## 3.4. ELC10 compared to existing maps

ELC10 produced by the final RF model compared favourably relative to two global and two European land cover products (Fig. 7). The overall accuracy for the ELC10 map was 18% higher than the lower resolution CORINE map, and 17% higher than the global 10-m FROM-GLC10 map. In comparison to the European-specific products, our map produced a 5% greater overall accuracy. Specifically, ELC10 was 7% more accurate than S2GLC and 3% more accurate than Pflugmacher et al. ELC10 displayed class-specific accuracies that were slightly (<1%) lower than Pflugmacher et al (2019) for wetland, bare land and cropland classes (Fig. 7). Otherwise, the ELC10 class-specific accuracies were greater than those for the other maps in all other land cover classes. Notable improvements upon other maps include those for water, and artificial land (Fig. 7).

In terms of the maps' utility for area statistics, the ELC10 map produced a strong correlation to official LUCAS-based statistics (high $R^2$ and low mean absolute error; Fig. 8e). Land cover class area estimates are within 4.19% of the observed value for ELC10. This error is marginally higher than the error from Pfludmacher et al (0.16% higher), but lower than the error for the other maps. Perhaps the most significant advantage of the ELC10 map is only realized at the landscape scale. Fig. 9 (and Fig. S3, 4 and 5) illustrates the ability of the ELC10 map to distinguish detailed landscape elements like hedge rows and intra-urban green spaces which are lost in the other lower-resolution products.

# 4. Discussion

## 4.1. Comparison to state of the art

The ELC10 map produced here has accuracy levels (90.2%) that are comparable with multiple city- and country-scale Sentinel-based land cover maps globally (Phiri et al., 2020). Within the European context, we find that ECL10 has 18% less error than the CORINE dataset which is widely used for research and accounting purposes. This corroborates results from others (Felicísimo and Sánchez Gago, 2002; Pflugmacher et al., 2019) who have also found uncertainty and bias associated with CORINE maps. The primary explanation for this discrepancy in accuracy is that the CORINE minimum mapping unit (25 ha) is very coarse compared to Landsat- and Sentinel-based maps (e.g. ELC10 minimum mapping unit of 0.01 ha). The CORINE project also adopts a bottom-up approach of consolidating nationally-produced land cover datasets into one and is therefore prone to inconsistencies and spatial variations in mapping error. Although CORINE has been used effectively to stratify probabilistic sampling of land cover for unbiased area estimates (Stehman, 2009), it may not be functional in small municipalities or for other land use and ecosystem models that require fine-grained spatial data.

To address the need for fine-grained land cover data, the European Space Agency recently initiated the development of the S2GLC map over Europe at 10m resolution (http://s2glc.cbk.waw.pl/) (Malinowski et al., 2020). The ELC10 map produced here extends on the S2GLC work by improving overall accuracy by 7% and adopting an open-source and transparent approach in a similar vein to the Landsat-based map by (Pflugmacher et al., 2019). Unlike previous pan-European maps, our approach relies on purely satellite-based input data and is therefore annually updatable for the foreseeable future lifespan of Sentinel and VIIRS sensors. It is thus independent of national topographic mapping datasets that take considerable resources to update (e.g. national land resource map of Norway; (Ahlstrøm et al., 2019)). ELC10 also leverages Google's cloud computing infrastructure, made freely available for research purposes through Google Earth Engine. We were able to train and make inference with our Random Forest model over 700 TB of satellite data at a rate of 100 000 km$^2$ per hour which equates to approx. 4 days of computing time to generate the 10-m product for Europe. In this way, regional or continental scale mapping of land cover, which has typically been the domain of large transnational institutions, may become more democratized and independent of political agendas (Nagaraj et al., 2020).

## 4.2. Potential applications

As satellite technology and cloud computing advances, the ability to map land cover at high spatial resolutions is becoming increasingly possible. This opens up a range of novel use-cases for land cover maps at continental scales. One example is for mapping small patches of green space within and outside of urban areas. (Rioux et al., 2019) found that urban green space cover and associated ecosystem services were generally underestimated at spatial resolutions coarser than 10 m. Similarly, green spaces constituting important habitat for biodiversity such as semi-natural grasslands are often not portrayed in current land cover maps. This is significant given that habitat loss is one of the main threats facing biodiversity, particularly pollinator species, in agricultural landscapes across Europe (Carvalheiro et al., 2013; Ridding et al., 2020). Quantifying and monitoring the remaining fragmented habitat is therefore a conservation concern at both regional and national levels (Janssen et al., 2016). This is also true for monitoring the corollary of habitat loss - habitat restoration initiatives. Agri-Environmental schemes (Cole et al., 2020) such as establishment of stone walls, hedge rows, and strips if semi-natural vegetation along field margins are not detected by current land cover mapping initiatives. High resolution land cover maps such as the ELC10, presented here, provide a means to monitor the status and trends of the remaining patches of semi-natural habitats and other small green spaces over Europe.

## 4.3. Limitations and opportunities

As with all land cover products, there are several limitations to ELC10 that are important to note in the interest of data users and future iterations of pan-European land cover maps. Our model produced classification errors that were greatest (accuracies below 80%) in southern Europe due to the predominance of, and spectral similarity between shrubland and bare land classes. For future refinements of the map one could aim to partition the LUCAS shrubland class into e.g. 2-3 levels of vegetational succession. Although some regions (i.e .central Europe, Fig 5) and classes (i.e. woodland: 95%, Table 2) exhibited much higher accuracies than southern Europe, the error rate may still be significant particularly in the context of monitoring land use changes. A 95% accuracy implies that a land cover class would have to change by 10% within a spatial unit (e.g. country or municipality) from year to year in order for a map like ELC10 to detect it with statistical confidence.

A major source of error in land cover models is the reference data. The LUCAS dataset is vulnerable to geolocation errors due to GPS malfunctioning in the field, interpretation errors, and land cover ambiguities. For instance, the European Environment Agency found that a post-screening of the LUCAS dataset increased CORINE-2000 accuracy by 6.4 percentage points (Büttner and Maucha, 2006). Apart from mis-labelled LUCAS points, intersecting Sentinel pixels may contain mixed land cover classes and therefore introduce noise into the spectral signal (d'Andrimont et al., 2020b). This is why the LUCAS Copernicus Module was initiated to produce quality-assured homogeneous polygons for integration with earth observation. However, here we found that intersecting Sentinel pixels with LUCAs polygon centroids did not significantly improve classification accuracy relative to the raw LUCAS point locations alone (Fig. 3e). This finding supports the well-established characteristic of Random Forest models which makes them robust against noisy training data (Pelletier et al., 2017).

Users of ELC10 should also be aware that our classification model is extrapolating into areas without any reference data in countries including Norway, Switzerland, Liechtenstein, and the non-EU Balkan states. However, because the LUCAS data covers a broad range of environmental conditions, it is reasonable to assume similar accuracies to neighboring countries, although this needs to be tested. The efficacy of integrating ground reference samples with remote sensing may be illustrative for Norway and other countries and stimulate future open-access land cover surveys. The fact that we found accuracies >85% with <5K reference points (Fig. 4) should act as encouragement because it shows that land cover mapping with earth observation does not necessarily require large resources dedicated to reference data collection. However, the variance in classification accuracy increases substantially with a reduction in reference sample size and therefore this might limit the ability to make accurate models at both national and continental scale. Alternatives to in-situ sampling include less resource-intensive methods, commonly adopted in the deforestation monitoring community, such as visual interpretation of very high resolution satellite or aerial imagery in platforms like Collect Earth Online (Saah et al., 2019).

There remain several avenues for improving upon ELC10 and Sentinel-based land cover mapping which may strengthen its utility for research and policy purposes. The harmonisation of Landsat and Sentinel time series (Shang and Zhu, 2019) may enhance the benefit gained from spectro-temporal features. This may be particularly beneficial in areas with high cloud cover which creates gaps in the Sentinel-2 time series and consequent noise in the spectro-temporal features. The use of repeat-pass SAR interferometry may also enhance accuracy (Sica et al., 2019) beyond those achieved here because we were limited to using Sentinel-1 Ground Range Detected data that is analysis-ready in Google Earth Engine. In this particular case, and in the inability to use products like "S1_GRD_FLOAT", Google Earth Engine is limited and one might explore other cloud computing platforms such as Sentinel Hub, Open Data Cube, or custom set-ups in Microsoft

Azure or Amazon Web Services (Gomes et al., 2020). Other cloud computing platforms may also offer a suite of other machine learning algorithms under the deep learning umbrella, such as neural networks, which may produce greater accuracies than classification tree approaches like RF (Ma et al., 2019). Although, recently Google Earth Engine has developed an integration with the machine learning platform TensorFlow which is allowing for the application of deep learning algorithms to land cover classification (e.g. Amani et al., 2020: Parente et al., 2020).

Finally, research on mapping uncertainty is an ongoing need. This is particularly true for quantifying uncertainty associated with land cover change statistics (Olofsson et al., 2013) derived from Sentinel land cover maps. Land cover change from Sentinel data may be assessed with the next iteration of LUCAS in 2021, or with the harmonized historical data provided by (d'Andrimont et al., 2020a). Quantifying uncertainty is necessary for such maps to be included into governmental and municipal accounting frameworks that ultimately contribute to addressing the global SDGs. Khatami et al., (2016) review a range of methods to derive pixel-level estimates of uncertainty, many of which rely on producing posteriori class probabilities obtained from the random forest classifier. Class probabilities may also be used to perform post-processing steps that remove artifacts and salt-and-pepper effects of pixel-based classification as in Malinowski et al., (2020).

### 4.4. Recommendations

We attempted to maintain transparency in the data pre-processing decisions we made by presenting the effects on model accuracy at each step (Fig. 2). Although our findings are not necessarily generalizable to areas outside of Europe, they are useful guidelines for others to learn from. Based on our experience, we recommend the following for future Sentinel-based land cover mapping at continental scales.

- The atmospheric correction of Sentinel-2 optical has marginal effects on classification accuracy and therefore may be skipped. This is supported by other studies (Rumora et al., 2020) and is particularly relevant when users are interested in near-real time land cover classification because Top of Atmosphere products are generally made available before Surface Reflectance products.
- Applying a speckle filter to Sentinel-1 imagery has marginal effects on classification accuracy and therefore may be skipped. As far as we are aware, there are no other studies that have tested this effect. Applying speckle filtering is computationally intensive and therefore excluding it benefit fast and on-the-fly land cover classifications where desirable. However, we acknowledge that we only used a single median and standard deviation per band and orbit mode for a full year of data. Speckle filtering may be more effective if one derives seasonal or monthly composites as inputs into the classifier as we did with Sentinel-2 NDVI.
- The fusion of Sentinel-1 and Sentinel-2 data has large increases on classification accuracy (3 - 10%) and is therefore encouraged. The addition of auxiliary variables that capture large-scale environmental gradients important for distinguishing spectrally similar classes (e.g. shrubland and forest) also improve classification accuracies and should be included. However, users should be cautious of spatial overfitting to these auxiliary variables which may cause geographical biases due to spatial autocorrelations (Meyer et al., 2019; Roberts et al., 2017).
- Cleaning reference samples through initiatives like the LUCAS Copernicus Module may not be worth the marginal gains in classification accuracy. RF models are robust against noisy training data (Pelletier et al., 2017) and therefore, so long as a clean validation sample is maintained, filtering noise in training data may not be necessary. Nevertheless, clean reference data supplied by the Copernicus Module is invaluable to deriving realistic accuracy estimates. We supplemented the Copernicus Module polygons with LUCAS points (n = 18 009) in order to balance class representativity in the training sample. We did this using an outlier removal procedure which may have artificially inflated our final accuracy estimates. Therefore, we recommend that initiatives like the Copernicus Module ensure that their sample is representative of the class area proportions in the study area so that augmenting the training sample is not necessary for earth observation applications in the future.

- Collecting tens of thousands of reference data points may also not be necessary depending on the desired classification accuracy. We find that accuracies above 85% are achievable with less than 5000 LUCAS points, albeit for an 8-class classification typology.
- Cloud computing infrastructure like Google Earth Engine make ideal platforms given that we could produce a pan-European map within approx. 4 days of computation time from a single research user account.

## 5. Conclusion

The recent proliferation of freely-available satellite data in combination with advances in machine learning and cloud computing has heralded a new age for land cover classification. What has previously been the domain of transnational institutions, such as the European Space Agency, is now open to individual researchers and members of the public. We present ELC10 as an open-source and reproducible land cover classification workflow that contributes to open science principles and democratizes large scale land cover monitoring. We find that combining Sentinel-2 and Sentinel-1 data is more important for classification accuracy than the atmospheric correction and speckle filtering pre-processing steps individually. We also confirm the findings of others that Random Forest is robust against noisy training data, and that investing resources in collecting tens of thousands of ground-truth points may not be worth the gains in accuracy. Despite the effects of data pre-processing, ELC10 has unique potential for quantifying and monitoring detailed landscape elements important to climate mitigation and biodiversity conservation such as urban green infrastructure and semi-natural grasslands. Looking to the future, maps like ELC10 can be annually updated, and repeated in-situ surveys like LUCAS can be used for quantifying uncertainty and accuracy in area change estimates. Quantifying uncertainty is crucial for earth observation products to be taken seriously by policy makers and land use planners.

## Acknowledgements

This work was supported by the Norwegian Research Council through the *Researcher Project for Young Talents* program (Grant number 302692).

## Data and code availability

The ELC10 dataset is available here: https://doi.org/10.5281/zenodo.4407051 JavaScript and R code to reproduce ELC10 is available here: https://github.com/NINAnor/ELC10

Table 1. Land cover typology adopted along with LUCAS codes and descriptions.

| Land cover label | LUCAS class definitions and sub-class inclusions and exclusions |
|---|---|
| Artificial land | Artificial land (A00): Areas characterized by an artificial and often impervious cover of constructions and pavement. Includes roofed built-up areas and non-built-up area features such as parking lots and yards. Excludes non-built-up linear features such as roads, and other artificial areas such as bridges and viaducts, mobile homes, solar panels, power plants, electrical substations, pipelines, water sewage plants, open dump sites. |
| Cropland | Cropland (B00): Areas where seasonal or perennial crops are planted and cultivated, including cereals, root crops, non-permanent industrial crops, dry pulses, vegetables, and flowers, fodder crops, fruit trees and other permanent crops. Excludes temporary grasslands which are artificial pastures that may only be planted for one year. |
| Woodland | Woodland (C00): Areas with a tree canopy cover of at least 10% including woody hedges and palm trees. Includes a range of coniferous and deciduous forest types. Excludes forest tree nurseries, young plantations or natural stands (< 10% canopy cover), dominated by shrubs or grass. |
| Shrubland | Shrubland (D00): Areas dominated (at least 10% of the surface) by shrubs and low woody plants normally not able to reach >5m of height. It may include sparsely occurring trees with a canopy below 10%. Excludes berries, vineyards and orchards. |
| Grassland | Grassland (E00): Land predominantly covered by communities of grassland, grass-like plants and forbs. This class includes permanent grassland and permanent pasture that is not part of a crop rotation (normally for 5 years or more). It may include sparsely occurring trees within a limit of a canopy below 10% and shrubs within a total limit of cover (including trees) of 20%. May include: dry grasslands; dry edaphic meadows; steppes with gramineae and artemisia; plain and mountainous grassland; wet grasslands; alpine and subalpine grasslands; saline grasslands; arctic meadows; set aside land within agricultural areas including unused land where revegetation is occurring; clear cuts within previously existing forests. Excludes spontaneously re-vegetated surfaces consisting of agricultural land which has not been cultivated this year or the years before; clear-cut forest areas; industrial "brownfields"; storage land. |
| Bare land | Bare land and lichens/moss (F00): Areas with no dominant vegetation cover on at least 90% of the area or areas covered by lichens/ moss. Excludes other bare soil, which includes bare arable land, temporarily unstocked areas within forests, burnt areas, secondary land cover for tracks and parking areas/yards. |
| Water | Water areas (G00): Inland or coastal areas without vegetation and covered by water and flooded surfaces, or likely to be so over a large part of the year. Also includes areas covered by glaciers or permanent snow |
| Wetland | Wetlands (H00): Wetlands located inland and having fresh water. Also wetlands located on marine coasts or having salty or brackish water, as well as areas of a marine origin. |

**Table 2.** Estimated error matrix for the final classification with estimates for user's accuracy (UA) and producer's accuracy (PA). Overall accuracy is 90.2%.

| | | Reference | | | | | | | | | | |
|---|---|---|---|---|---|---|---|---|---|---|---|---|
| | Prediction | 1 | 2 | 3 | 4 | 5 | 6 | 7 | 8 | Total | UA (%) | SE |
| 1 | Artificial land | 2339 | 57 | 8 | 22 | 3 | 0 | 0 | 4 | 2433 | 96.1 | 0.4 |
| 2 | Bare land | 15 | 1219 | 5 | 43 | 54 | 19 | 7 | 17 | 1379 | 88.4 | 0.8 |
| 3 | Cropland | 13 | 124 | 16251 | 931 | 190 | 0 | 11 | 172 | 17692 | 91.9 | 0.2 |
| 4 | Grassland | 19 | 118 | 1171 | 13378 | 499 | 5 | 62 | 442 | 15694 | 85.2 | 0.3 |
| 5 | Shrubland | 6 | 120 | 207 | 255 | 3002 | 0 | 5 | 404 | 3999 | 75.1 | 0.7 |
| 6 | Water | 0 | 20 | 1 | 5 | 0 | 1110 | 15 | 2 | 1153 | 96.3 | 0.5 |
| 7 | Wetland | 0 | 48 | 11 | 28 | 24 | 2 | 2379 | 59 | 2551 | 93.3 | 0.5 |
| 8 | Woodland | 6 | 126 | 280 | 502 | 719 | 2 | 23 | 23288 | 24946 | 93.4 | 0.2 |
| | Total | 2398 | 1832 | 17934 | 15164 | 4491 | 1138 | 2502 | 24388 | 69847 | | |
| | PA (%) | 97.5 | 66.5 | 90.6 | 88.2 | 66.8 | 97.5 | 95.1 | 95.5 | | 90.2 | |
| | SE | 0.9 | 0.6 | 0.2 | 0.3 | 0.7 | 0.3 | 0.7 | 0.1 | | | 0.1 |

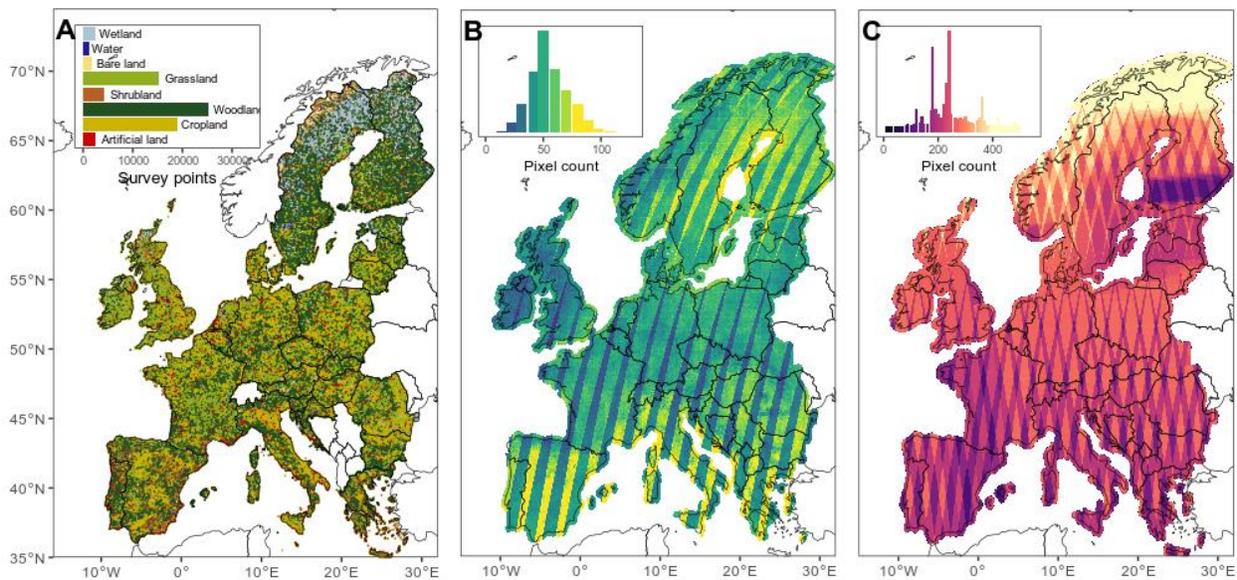

**Figure 1.** Study area with available land cover reference points (A) and Sentinel (B, C) satellite imagery. Each point in A is a sampling location (53 476 polygons and 282 854 points) with a land cover class label. The number of available cloud-free Sentinel-2 pixels and Sentinel-1 pixels during 2018 are mapped in B and C, respectively.

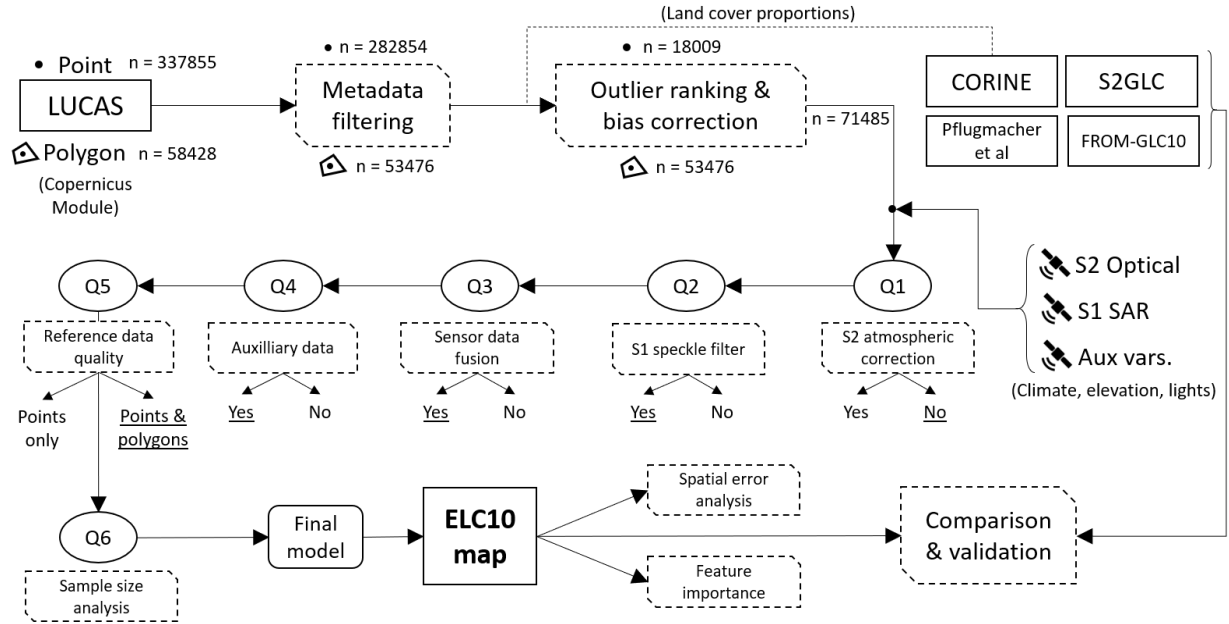

**Figure 2.** Methodological workflow for evaluating pre-processing decisions in generating the final ELC10 land cover map. Underlined outcomes are those that were chosen for the final model. Abbreviations: S1 - sentinel 1; S2 - sentinel 2; Aux vars - auxiliary variables.

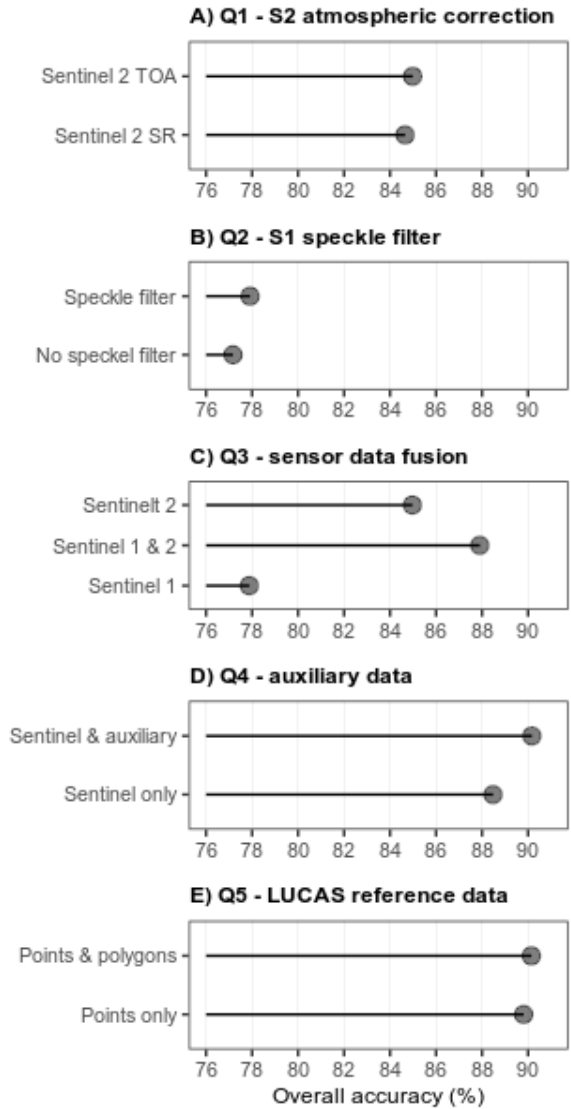

**Figure 3.** The effect of pre-processing decisions on land cover classification accuracy. Random Forest model overall accuracies are displayed for alternative Sentinel 2 (A), and 1 (B) pre-processing steps, Sentinel 1 and 2 data fusion options (C), the addition of auxiliary variables (D), and the quality of reference data (E).

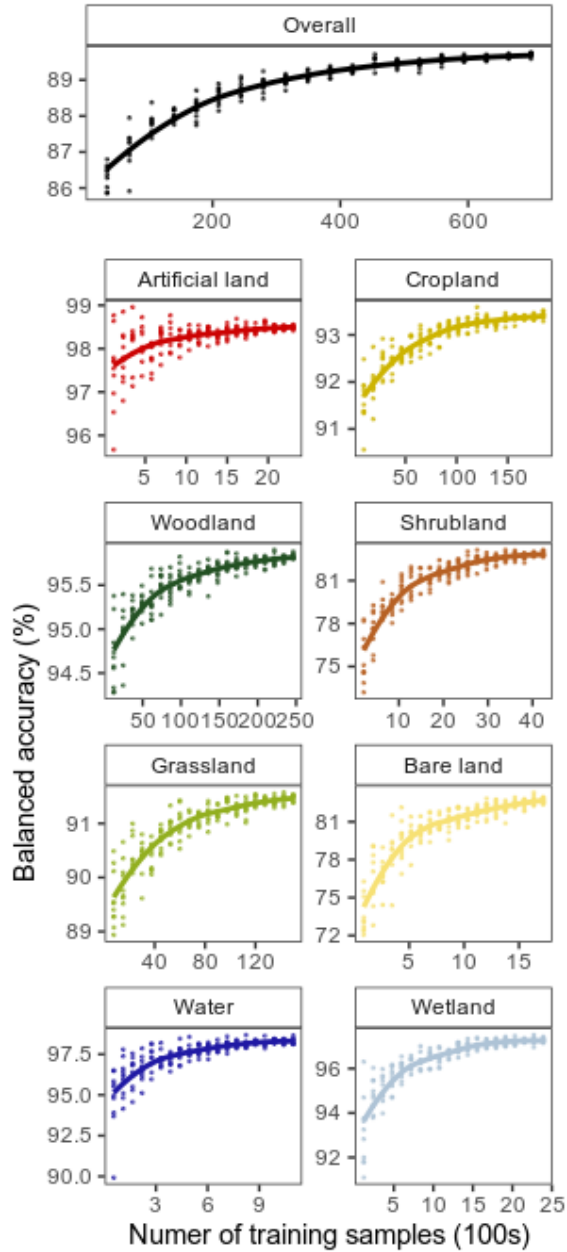

**Figure 4.** The effect of reference sample size on overall and class-specific accuracy. Random Forest classification models were trained on iteratively smaller sample sizes. Points in each facet plot represent bootstrapped (n = 10) model accuracy estimates and are fit with Loess regression lines.

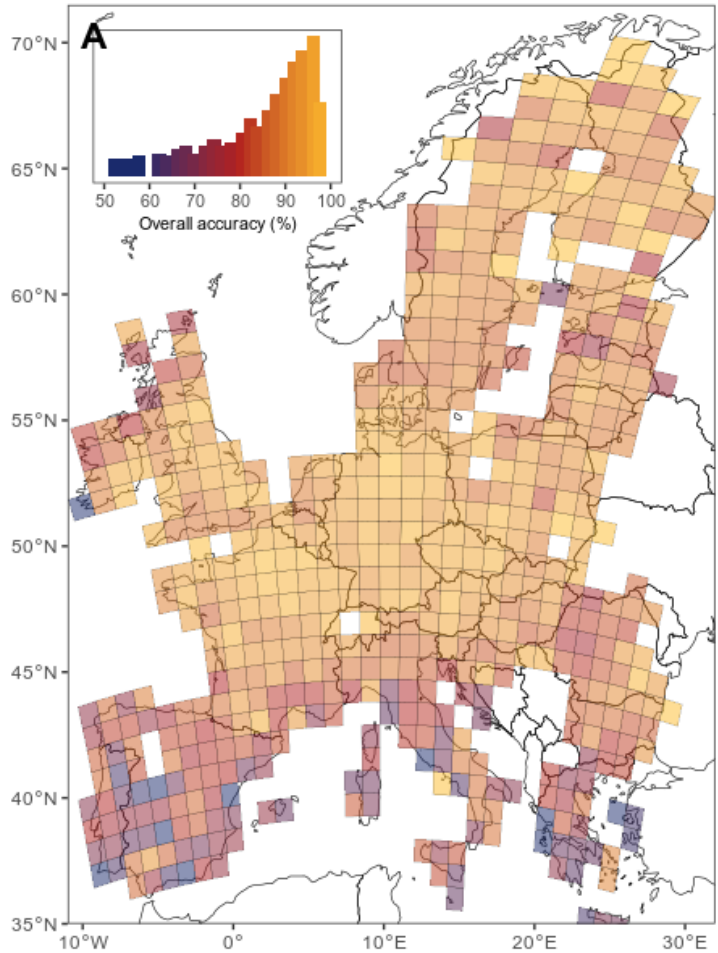

**Figure 5.** Map showing land cover classification accuracy over 100x100km grid squares. The inset bar plot shows the abundance of grid squares across the range of error (percentage overall accuracy). Missing grid cells are where there were insufficient validation samples to construct an error matrix.

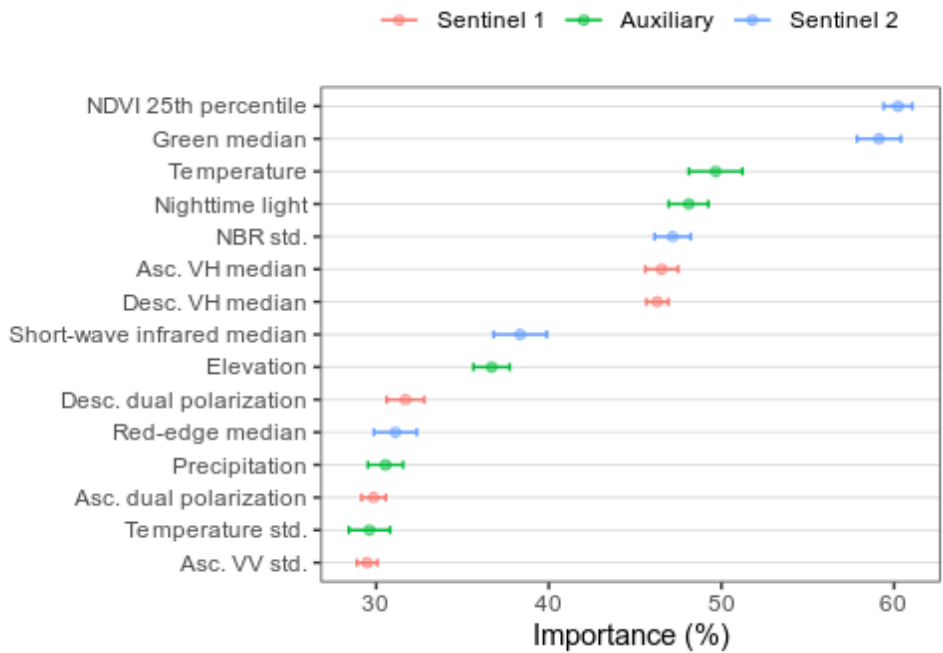

**Figure 6**. Variable importance plot showing the relative contribution of the top 15 most influential features in the final Random Forest classification model. Points and error bars represent means and standard errors based on bootstrapped (n = 10) model runs.

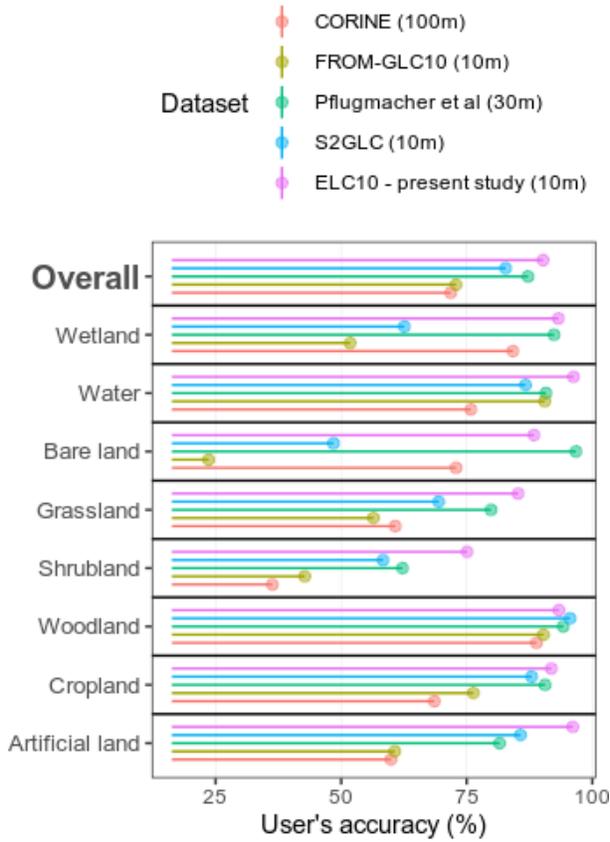

**Figure 7.** Class-wise user's and overall accuracy for different European land cover products. Horizontal lines and points show the accuracy achieved for each land cover map.

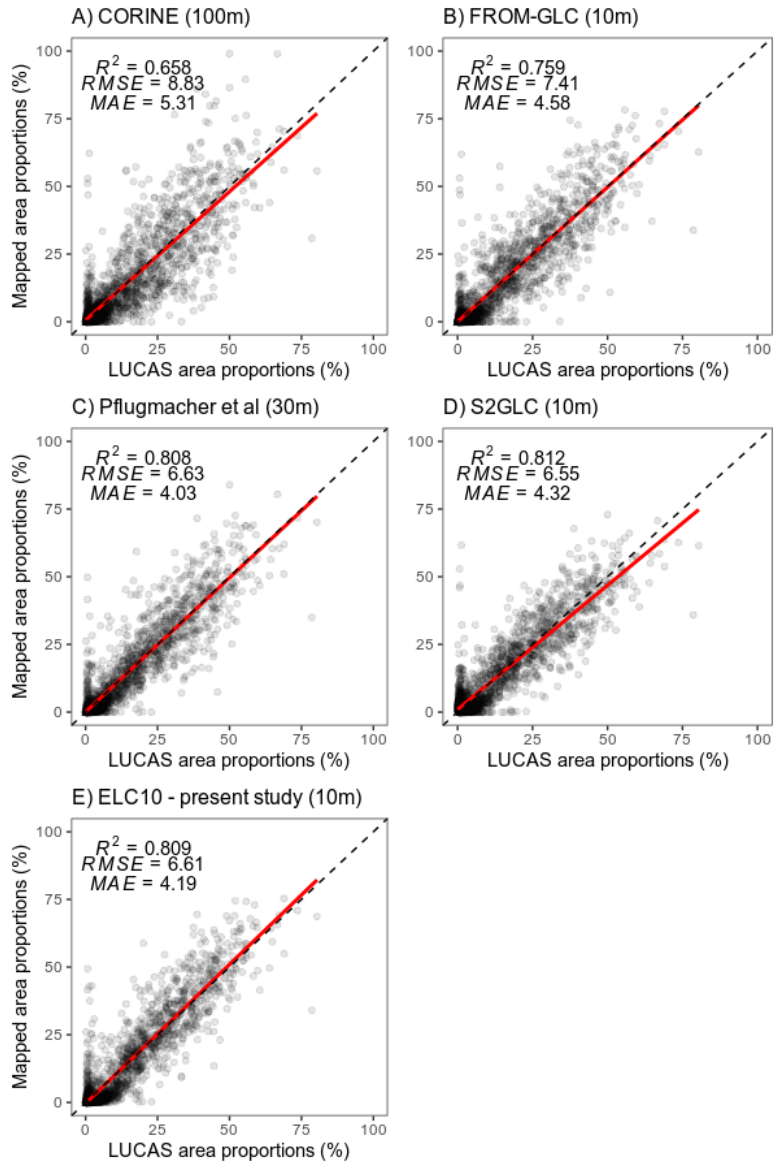

**Figure 8.** Correlation of mapped land cover proportions with LUCAS accounting statistics for each European land cover product. Each data point represents the proportion for a NUTS level 2 statistical unit. Coloured linear regression lines are fitted per land cover class with an overall regression in black. Overall regression $R^2$, root mean square error (*RMSE*) and mean absolute error (*MAE*) are reported.

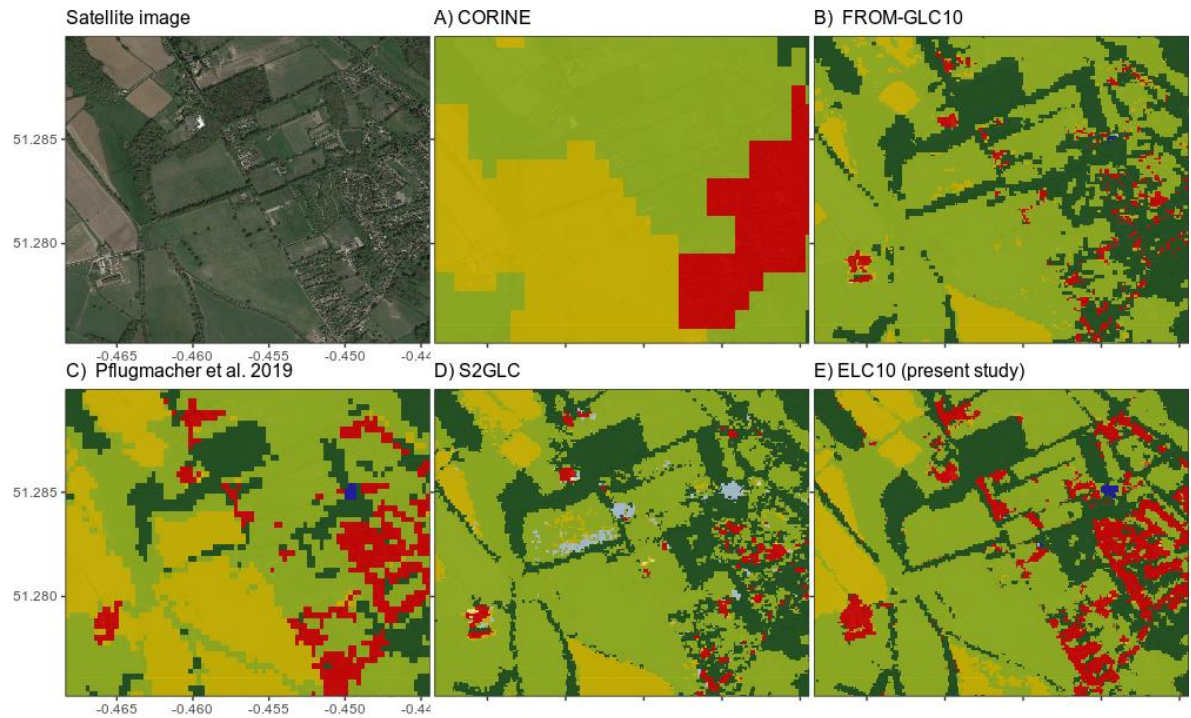

**Figure 9.** Example of land cover classifications at the local scale for a selected landscape in Woking (south of London, England). Maps are shown for the present study relative to the four comparative datasets. Please refer to supplement Fig. S3, S4 and S5 for more comparative examples.

# Supplementary tables and figures

**Table S1.** Land cover maps used for comparison with ELC10 were relassified into the ELC10 (based on LUCAS high level typology) typology. The lookup tables to show reclassifications are presented below.

| FROM-GLC10 | ELC10 |
| --- | --- |
| Cropland | Cropland |
| Forest | Woodland |
| Grassland | Grassland |
| Shrubland | Shrubland |
| Wetland | Wetland |
| Water | Water |
| Tundra | Grassland |
| Artificial | Artificial land |
| Bare land | Bare land |
| Snow/ice | Water |

| S2GLC | ELC10 |
| --- | --- |
| Artificial surfaces | Artificial land |
| Cultivated areas | Cropland |
| Vineyards | Cropland |
| Broadleaf tree cover | Woodland |
| Coniferous tree cover | Woodland |
| Herbaceous vegetation | Grassland |
| Moors and heathlands | Grassland |
| Sclerophyllous vegetation | Shrubland |
| Marshes | Wetland |
| Peatbogs | Wetland |
| Natural material surfaces | Bare land |
| Permanent snow covered surfaces | Water |
| Water bodies | Water |

| Pflugmacher | ELC10 |
| --- | --- |
| Artificial land | Artificial land |
| Cropland seasonal | Cropland |
| Cropland perennial | Cropland |
| Forest broadleaf | Woodland |
| Forest coniferous | Woodland |
| Forest mixed | Woodland |
| Shrubland | Shrubland |
| Grassland | Grassland |
| Bare land | Bare land |
| Water | Water |
| Wetland | Wetland |
| Snow/ice | Water |

| CORINE | ELC10 |
|---|---|
| Urban fabric | Artificial land |
| Industrial, commercial, and transport units | Artificial land |
| Mine, dump, and construction sites | Bare land |
| Artificial, non-agricultural vegetated areas | Artificial land |
| Arable land | Cropland |
| Permanent crops | Cropland |
| Pastures | Grassland |
| Heterogeneous agricultural areas | Cropland |
| Forests | Woodland |
| Scrub and/or herbaceous vegetation associations > Natural grasslands | Grassland |
| Scrub and/or herbaceous vegetation associations > Moors and heathland, Sclerophyllous vegetation, Transitional woodland-shrub | Shrubland |
| Open spaces with little or no vegetation | Bare land |
| Inland wetlands | Wetland |
| Maritime wetlands | Wetland |
| Inland waters | Water |
| Marine waters | Water |

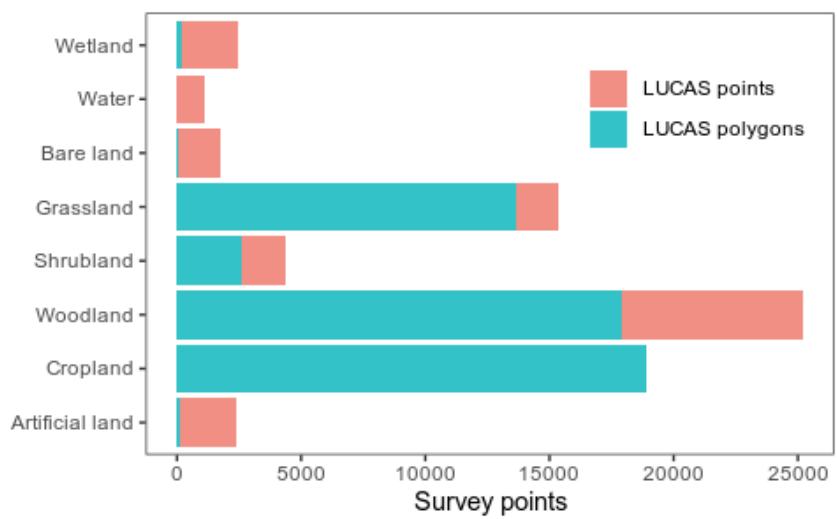

**Figure S1.** Distribution of LUCAS reference points used in the final ELC10 model (n = 71 485) across land cover classes. LUCAS polygons were supplemented with LUCAS points so that the samples sizes were proportional to the CORINE land cover proportions over Europe.

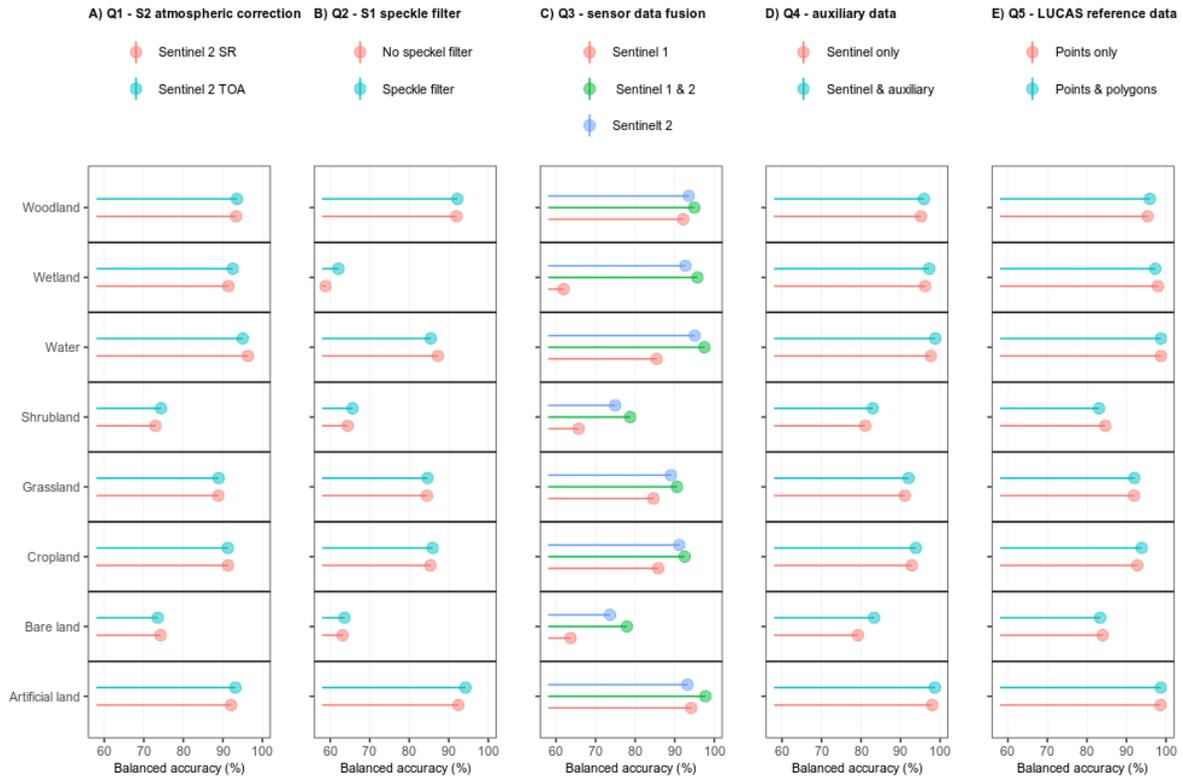

**Figure S2.** The effect of pre-processing decisions on land cover classification accuracy per land cover class. Random Forest model class-specific balanced accuracies are displayed for alternative Sentinel 2 (A), and 1 (B) pre-processing steps, Sentinel 1 and 2 data fusion options (C), the addition of auxiliary variables (D), and the quality of reference data (E).

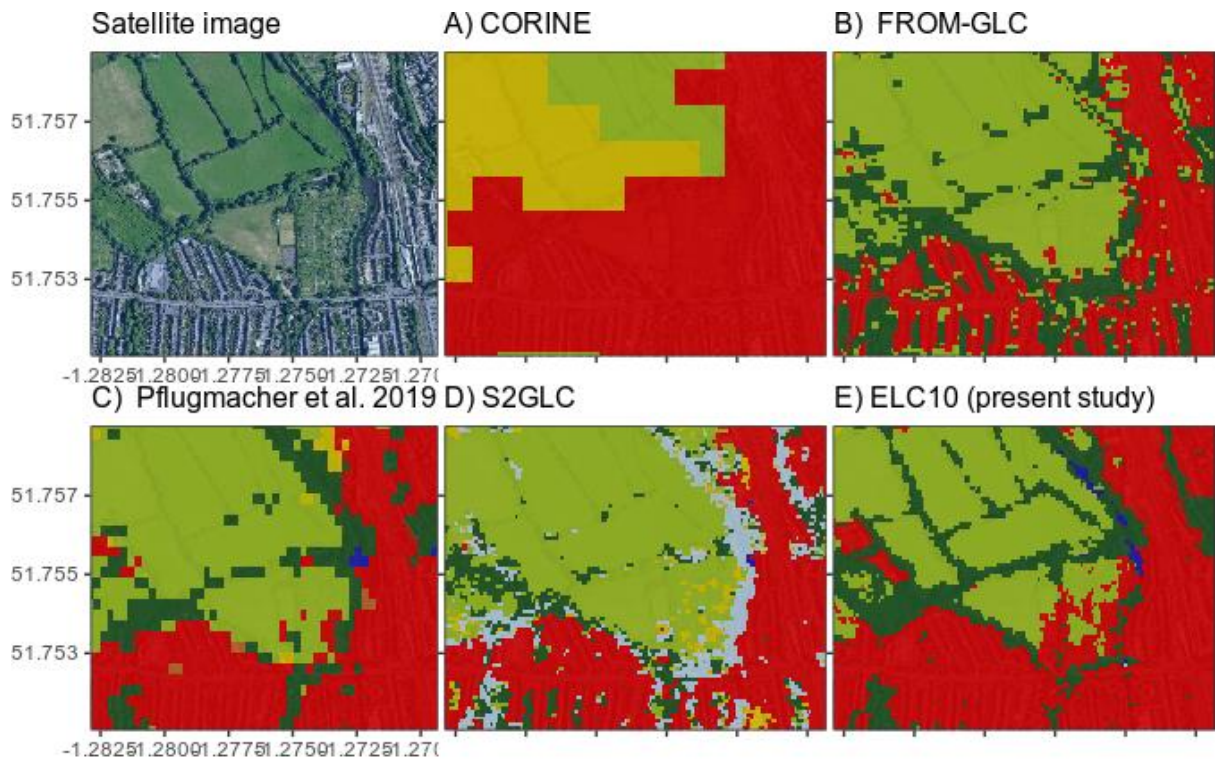

**Figure S3.** Example of land cover classifications at the local scale for a selected landscape in Ozford, England. Maps are shown for the present study relative to the four comparative datasets.

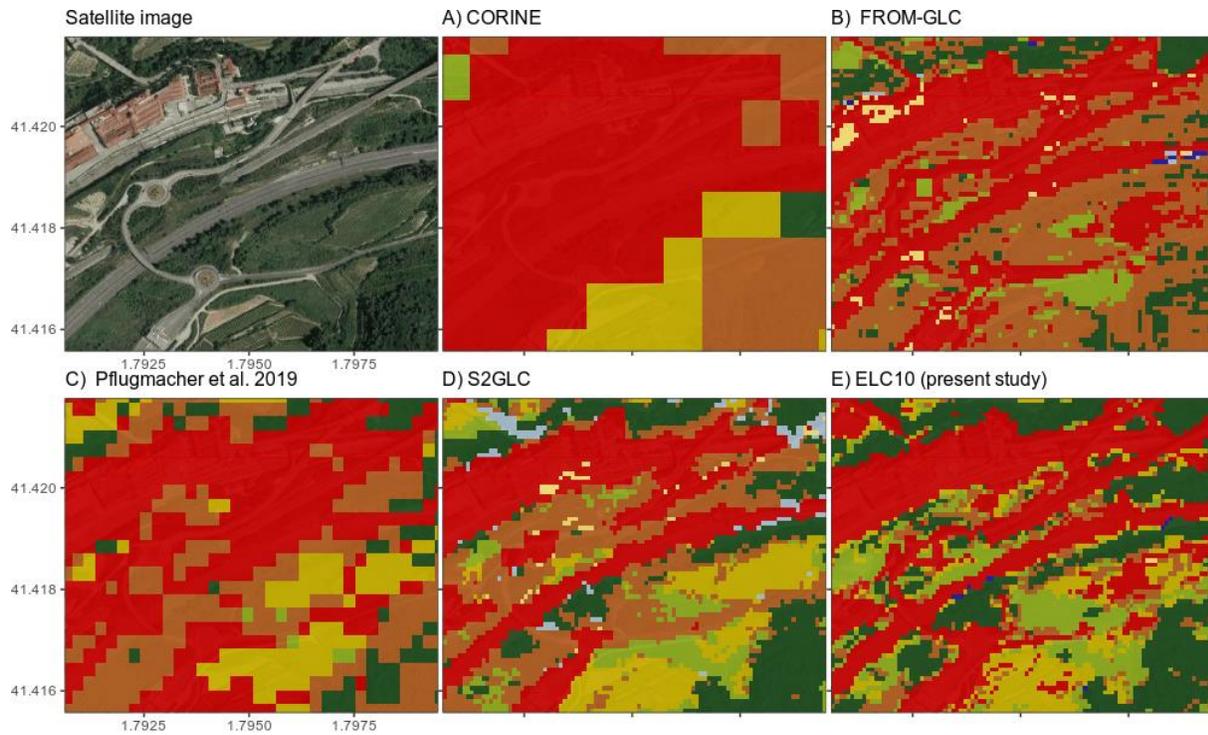

**Figure S4.** Example of land cover classifications at the local scale for a selected landscape east of Barcelona, Spain. Maps are shown for the present study relative to the four comparative datasets.

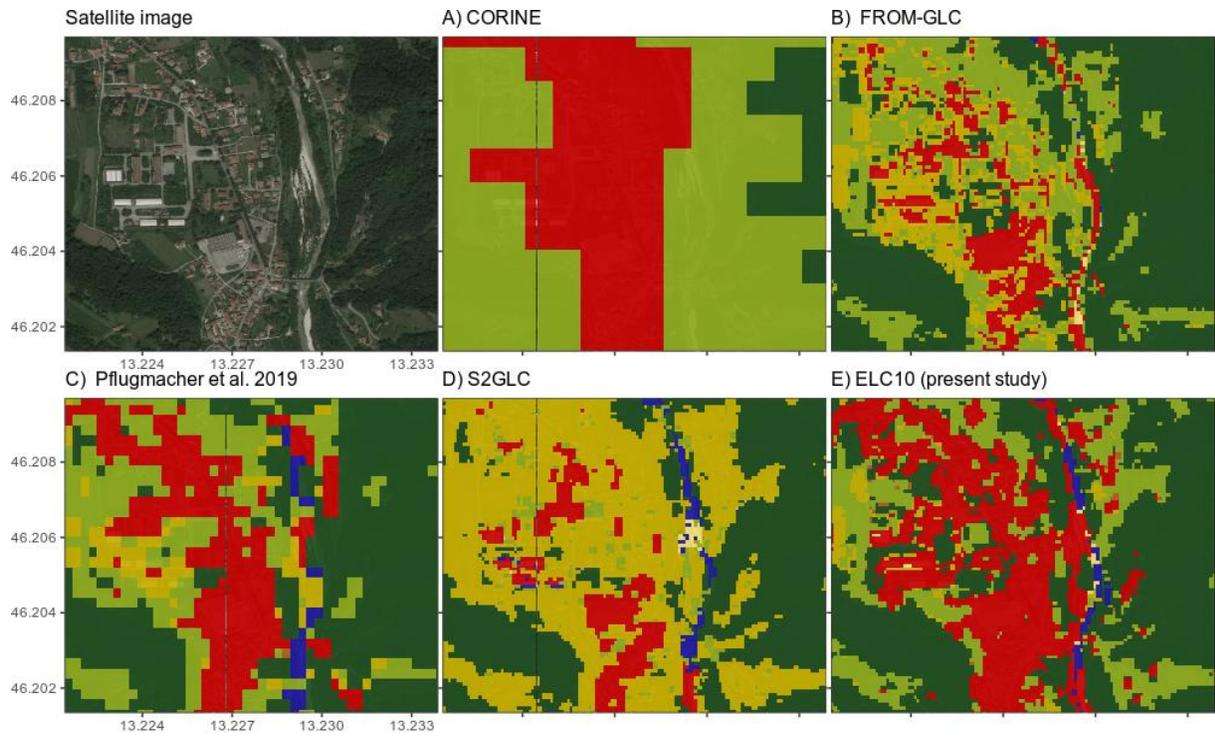

**Figure S5.** Example of land cover classifications at the local scale for a selected landscape south of Tarcento, Italy. Maps are shown for the present study relative to the four comparative datasets.